\documentclass[fleqn]{article}
\usepackage[utf8]{inputenc}
\usepackage{appendix}
\usepackage[hidelinks]{hyperref}
\usepackage{mathtools}
\usepackage{amsmath}
\usepackage{tikz}
\usepackage{mathdots}
\usepackage{yhmath}
\usepackage{cancel}
\usepackage{color}
\usepackage{siunitx}
\usepackage{array}
\usepackage{multirow}
\usepackage{amssymb}
\usepackage{gensymb}
\usepackage{tabularx}
\usepackage{booktabs}
\usepackage{svg}
\usepackage{pdfpages}
\usepackage{comment}
\usepackage{authblk}
\usepackage{subcaption, graphicx, float}
\usepackage{authblk}

\usepackage[backend=bibtex, style=vancouver, sorting=none]{biblatex} 
\usetikzlibrary{fadings}
\usetikzlibrary{patterns}
\usetikzlibrary{shadows.blur}

\title{Neural networks for dengue forecasting: \\ a systematic review}

\author[1]{L. Lober}
    
\affil[1]{%
    Departamento de Matemática Aplicada e Estatística, Instituto de Ciências Matemáticas e de Computação, Universidade de São Paulo—Campus de São Carlos, Caixa Postal 668, 13560-970 São Carlos, São Paulo, Brazil
}%

\author[1]{F. A. Rodrigues}

\author[1,2]{K. O. Roster}%
\affil[2]{%
    Harvard T. H. Chan School of Public Health, Boston, MA.
}%
    
\date{\today}

\begin{document}
\maketitle

\begin{abstract}

     \textbf{Background}: Early forecasts of dengue are an important tool for disease mitigation. Neural networks are powerful predictive models that have made contributions to many areas of public health. In this study, we reviewed the application of neural networks in the dengue forecasting literature, with the objective of informing model design for future work.\\
     \textbf{Methods}: Following PRISMA guidelines, we conducted a systematic search of studies that use neural networks to forecast dengue in human populations. We summarized the relative performance of neural networks and comparator models, architectures and hyper-parameters, choices of input features, geographic spread, and model transparency.\\
     \textbf{Results}: Sixty two papers were included. Most studies implemented shallow feed-forward neural networks, using historical dengue incidence and climate variables. Prediction horizons varied greatly, as did the model selection and evaluation approach. Building on the strengths of neural networks, most studies used granular observations at the city level, or on its subdivisions, while also commonly employing weekly data. Performance of neural networks relative to comparators, such as tree-based supervised models, varied across study contexts, and we found that 63\% of all studies do include at least one such model as a baseline, and in those cases about half of the studies report neural networks as the best performing model.\\
     \textbf{Conclusions}: The studies suggest that neural networks can provide competitive forecasts for dengue, and can reliably be included in the set of candidate models for future dengue prediction efforts. The use of deep networks is relatively unexplored but offers promising avenues for further research, as does the use of a broader set of input features and prediction in light of structural changes in the data generation mechanism.
\end{abstract}

\section{\label{sec:intro}Introduction}

Dengue fever is a mosquito-borne disease with a significant global burden. More than 100 countries are at risk of infection \cite{WHO2024}, with a substantial increase registered in the last five years, including European and eastern Mediterranean regions \cite{ECDC2025}. 2025, up to July, accounted for 3.6 million symptomatic dengue fever cases worldwide, a number that is possibly under-reported considering evidence from previous years \cite{WHO2024}. 

Dengue fever presents as a flu-like illness with symptoms ranging from mild to severe, with potentially life-threatening complications arising from severe dengue. The virus occurs in four serotypes, meaning up to four infections are possible for each person over their lifetime, and no specific treatment currently exists.

Preventing the circulation of the arboviruses and early warnings to public health agencies, including vector control methods and pinpointing hotspots of the disease, are essential methods for dengue mitigation, even considering recently available vaccination strategies against all dengue serotypes, which are already available in multiple countries \cite{WHO2024, Takeda2025, Ranzani2025Aug}. To that end, predictive models can help efficiently allocate public health resources to combat dengue, with the goal of reducing the overall disease burden. 

Dengue incidence is associated with many different risk factors. These include climate conditions such as rainfall, extreme weather events, and temperature \cite{Chien2014Dec, Thu1998Jun, Fan2014Mar}, land use \cite{Kilpatrick2012Dec}, and poverty \cite{Mulligan2015Feb}. These factors, and others related to dengue, may have nonlinear, context-specific, and time-varying effects on disease incidence, which poses a challenge to disease modeling. 

The literature on dengue forecasting is multi-disciplinary, uniting expertise from areas such as epidemiology, environmental science, computer science, and mathematics. Modeling frameworks include both theoretical and data-driven approaches, spanning dynamic transmission models, statistical time series forecasting, and machine learning methods. 

Machine learning models benefit from the increasing data availability on risk factors on the spreading of diseases, and offer non-parametric approaches that require less detailed knowledge of the disease and its context \cite{Guo2017Oct}. Neural networks are a subset of these algorithms, which have made significant contributions to medicine and public health, including applications such as medical image analysis for disease diagnosis \cite{Mienye2025Mar}, identifying abnormalities in signals such as electrocardiographs (ECG) \cite{Haglin2019Jan}, and optimizing decisions of health care providers, hospitals, and policy-makers \cite{Shahid2019Feb}. 
Neural networks are also used to forecast diseases, including malaria \cite{Zinszere2012}, influenza \cite{Alessa2018Dec}, and COVID-19 \cite{Bullock2020Nov}. 

This review examined the use of neural networks for dengue fever prediction, focusing specifically on:

    \begin{itemize}
        \item the technical decisions made in the literature, including architecture selection, hyper-parameter tuning and model proposals;
        \item input data, such as climate and demographic variables;
        \item the relative performance of different neural network architectures and comparator models, such as other machine learning techniques;
        \item geographic and temporal evolution of publications.
    \end{itemize}

\section{Methodology}

This review followed the Preferred Reporting Items for Systematic Reviews and Meta-Analyses (PRISMA) statement guidelines \cite{prisma2021}. The PRISMA flow chart for this study is represented in figure \ref{fig:prisma flowchart}. A systematic search was conducted using Web of Science/Knowledge, Scopus (abstract, title, keywords), PubMed, and Science Direct (abstract, title) databases. References of papers appearing in the search results were also examined for relevant works. The searches were conducted in September 10 2024, and used the following search string:
\begin{center}
    ``(deep learning OR neural network) AND dengue"
\end{center}

The papers were examined for inclusion in the review in three stages: by title, by abstract, and finally by full text. In each phase, the following inclusion criteria were applied conservatively. If the information in the title and/or abstract was inconclusive, the paper was included in the full-text review. These inclusion criteria were as follows:

    \begin{itemize}
        \item Studies must implement a neural network or deep learning technique, either as the main method or as a comparator model. Reviews of the literature on neural networks for dengue forecasting were also considered; 
        \item Studies must predict dengue fever incidence or risk. There were no restrictions on the type of target variable used. For example, the number of dengue cases, dengue incidence rate, or a binary dengue risk variable were all accepted as target features;
        \item Studies must examine dengue in a human population. Models for disease diagnosis of individuals were excluded. Studies modeling the location of vectors without relation to dengue incidence were excluded. Studies examining animal hosts were excluded;
        \item As the search query is in English, only English language articles were identified and included in the review.
    \end{itemize}

Alongside these criteria, two other standards were used when screening the full texts of selected papers: 

\begin{itemize}
    \item Data sources and processing steps must have been clearly stated;
    \item Methodology must be clear and well structured, and there was no apparent data leakage in its proceedings. Metrics and models, along with all relevant parameters to the chosen NN architecture, when available, were specified.
\end{itemize}

We also excluded papers that did not allow institutional access to the authors for reviewing, which was the case for three studies that fulfilled the eligibility criteria up to the full-text screening stage.

The results of each stage of the screening process can be seen in \autoref{fig:prisma flowchart} on \autoref{sec: appendix prisma}.


\section{Results}

\subsection{Data Sources}

    \begin{figure}[!h]
        \centering
        \includegraphics[width=0.9\textwidth]{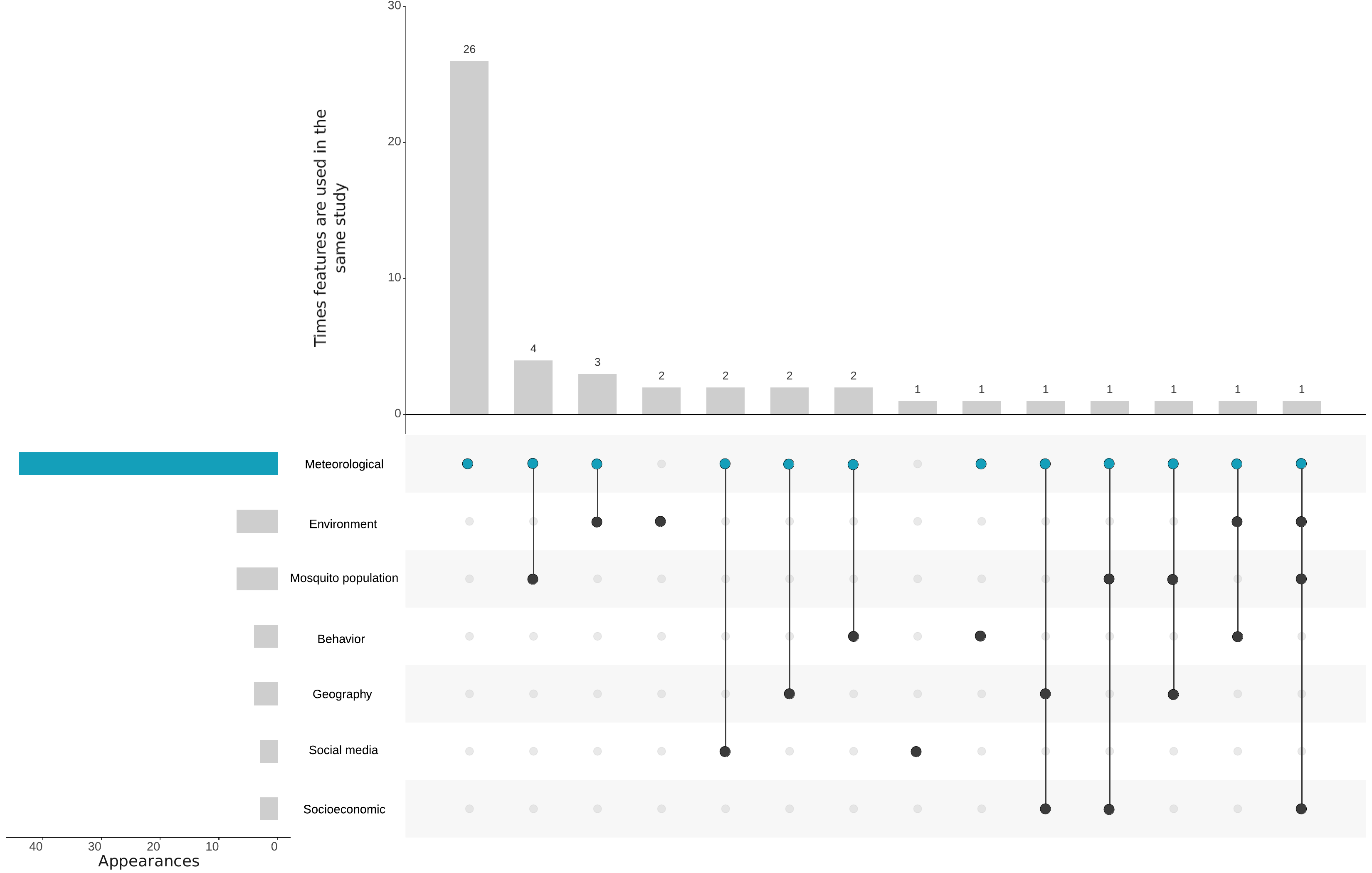}
        \caption{Auxiliary variables to dengue epidemiological data included in the selected studies, displaying the frequency with which studies included various categories of predictor variables and their combinations. The top bars measure their intersection size in the selected papers, and the central table shows the interplay between multiple variables. 'Meteorologic' data is present in most (69\%) studies.} 
        \label{fig:variables}
    \end{figure}

Meteorologic features are the most common data source, being featured in 43 of the 61 papers included in this study. This data comprises temperature, humidity, wind speed, solar radiation, pressure, rainfall, sea surface temperature measurements and El Niño \cite{tab56_Tran} data.

Out of all papers that did not include meteorologic data, only four \cite{tab11_Anderson2018, tab43_Andersson2019, tab45_Livelo2018, tab60_Araujo} add any extra features to train the neural networks beyond the epidemiologic information on dengue fever. In \cite{tab11_Anderson2018, tab43_Andersson2019}, both authors used environmental data coming from satellite or street imaging in conjunction to epidemiologic data for their forecasts; \cite{tab45_Livelo2018} instead uses social media data to complement their models. \cite{tab60_Araujo} on the other hand employs data on bus lines to understand how mobility restrictions in the pandemic years impacted dengue transmission in urban areas.

The second most frequent data source used for dengue fever forecasting was environmental data, found both as satellite images \cite{tab47_Rehman_2019}, vegetation indexes \cite{tab30_Dharmawardana2017, tab33_Li2022, tab35_Anggraeni2024, tab58_Hu} and street imaging from the Google StreetView program \cite{tab11_Anderson2018, tab43_Andersson2019}.

Following closely the number of uses of environment information, data on mosquitoes for forecasting dengue consisted on mosquito density \cite{tab35_Anggraeni2024} and density per person in a given month \cite{tab32_Dinh2016}, larvae free factor \cite{tab38_Anggraeni2022}, biting rate \cite{tab59_Lu}, entomological data in the form of raw number of \textit{Ae. Aegypti} and \textit{Ae. Albopictus} mosquitoes \cite{tab39_Dong2022}, and also the infection rate of larvae, male and female mosquitoes \cite{tab50_Kesorn2015}. Moreover, \cite{tab58_Hu} used species distribution model for dengue transmitting mosquitoes and \textit{Ae. Aegypt} temperature suitability.

Other, rarely-occurring features, found throughout the studies were human behavior, social media, geographic and socioeconomic data, which includes:

    \begin{itemize}
        \item \textbf{Behavior}: number of holidays in a given week, used as a boolean variable in \cite{tab26_Bergero2023}, human mobility through call data records (CDR) processing in \cite{tab30_Dharmawardana2017}  bus lines in \cite{tab60_Araujo} and time spent in transit stations from Google Regional Mobility Reports \cite{tab63_Roster2024}.
        
        \item \textbf{Social media}: short messages from X (formerly Twitter) containing dengue-related information, selected through preprocessing methodologies to ascertain the relevance of the social media data to dengue circulation \cite{tab5_Mussumeci2020, tab45_Livelo2018, tab62_Salsabiila}.
        
        \item \textbf{Geography}: \cite{tab10_Anno2024} included this information from data on the place of residence of infected individuals, labeled as township or village. We also listed population density from \cite{tab50_Kesorn2015, tab57_Das} considering land use, explicitly used in \cite{tab61_Francisco}, to dengue forecast;
        
        \item \textbf{Socioeconomic data}: \cite{tab39_Dong2022} used several extra information to improve their dengue forecasts: literacy, access to health services, and housing and sanitation information, which included houses with dirt floors, toilet facility, water pipelines, sewage system and electricity.
    \end{itemize}

Lastly, epidemiologic data, e.g. case counts or incidence rates, are necessarily included in all studies due to the nature of the prediction problem.
    
    \begin{figure}[!h]
        \centering
        \begin{minipage}{.42\textwidth}
            \includegraphics[width=0.9\textwidth]{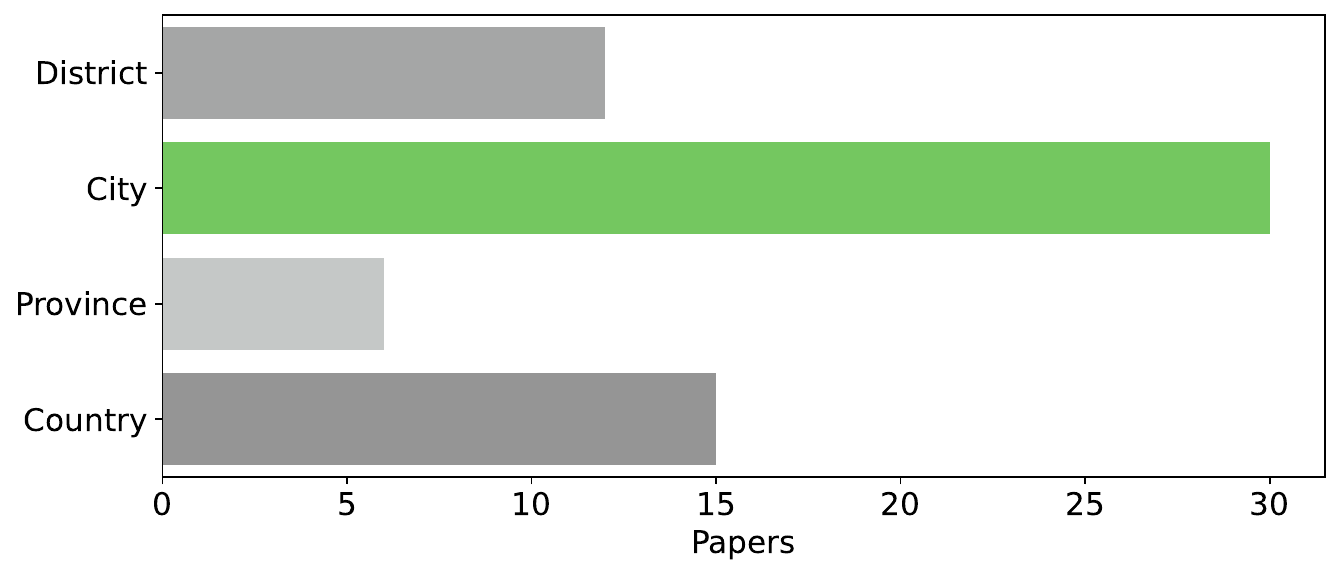}
        \end{minipage}
        \begin{minipage}{.42\textwidth}
            \includegraphics[width=0.9\textwidth]{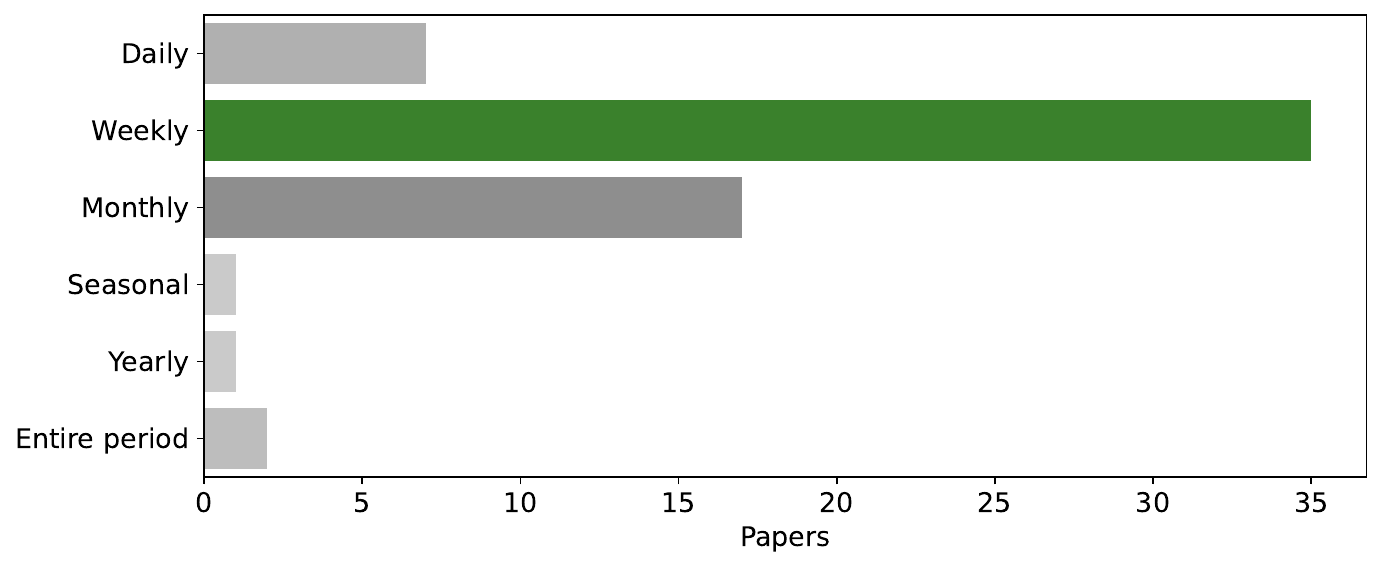}
        \end{minipage}
        \caption{Geographic and temporal distribution, with green bars highlighting the most frequent category in each case. The left panel shows the frequency with which studies predict dengue cases at the city, district, province, and country level. The right panel shows the frequency with which studies predict dengue cases at daily, weekly, monthly, seasonal, and yearly level. The ``entire period" of dengue data accounts for total number of cases in a certain period, and was used by studies that predicted dengue fever risk in specific regions \cite{tab11_Anderson2018, tab43_Andersson2019}.}
        \label{fig:granularity}
    \end{figure}

Studies focused both on macro regions or smaller divisions of a given area for their forecasts, covering forecasts at district-, city-, province-, and country-level (\autoref{fig:granularity}). ``Province" stands for a subdivision of countries, such as states, and ``district" are subdivisions of cities, which included both neighborhoods and the health center coverage area within a city.

    \begin{figure*}[!h]
        \centering
            \includegraphics[width=0.9\linewidth]{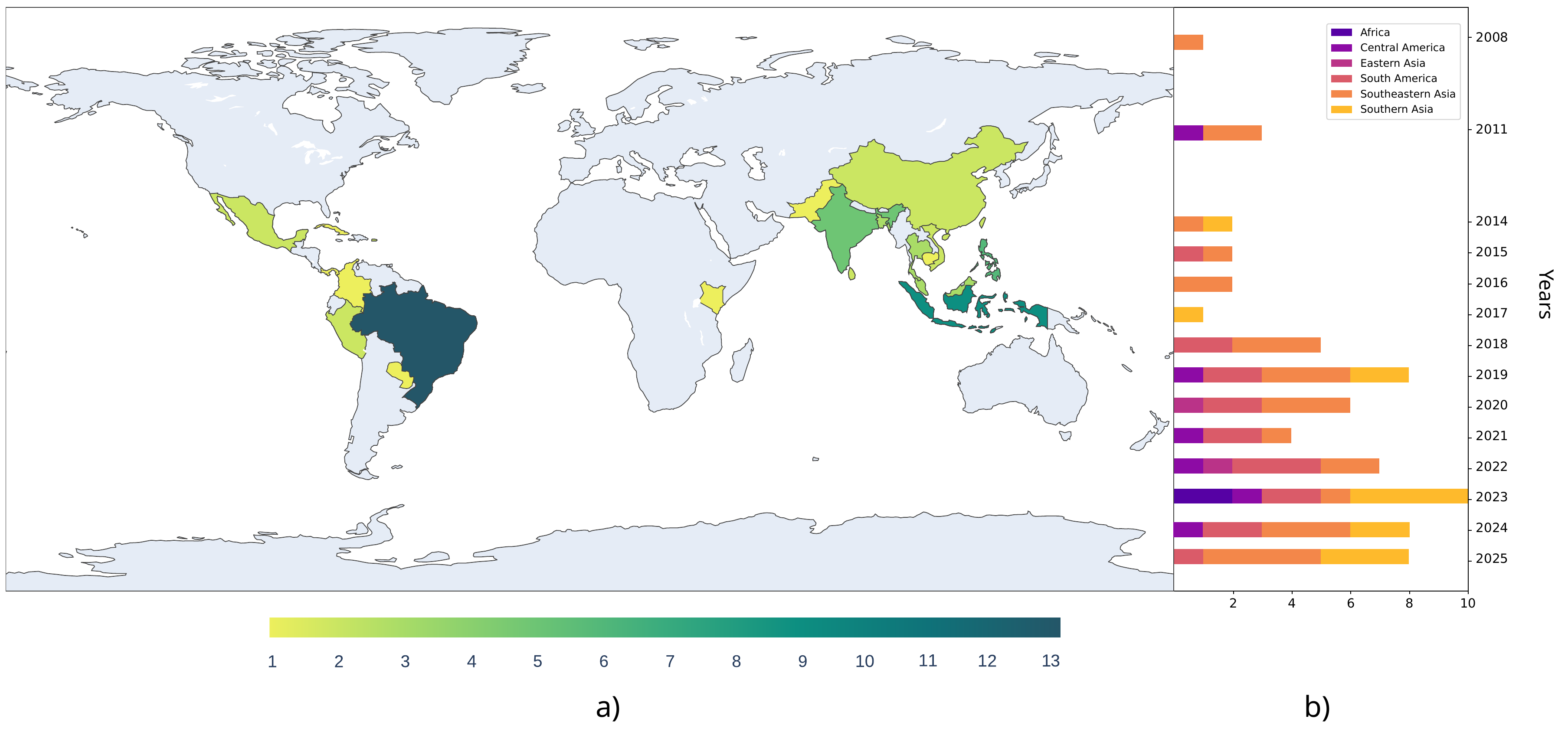}
        \caption{Geographic coverage. a) Number of studies predicting dengue in each country from 2008 to 2025 and b) number of studies predicting dengue in different geographic regions, by year of publication.}
        \label{fig:countries map}
    \end{figure*}


In \autoref{fig:countries map}, Brazil was the most frequent country for studies forecasting dengue fever, with 12 papers using Brazilian data in their analyzes, followed by Indonesia (8 papers) and India (5 papers). There was an increase in publication of studies using neural networks to forecast dengue starting in 2017, with 2019, 2025 and 2023, in this order, featuring the highest number of papers using this methodology.

\begin{figure}[!h]
    \centering
    \includegraphics[width=0.8\linewidth]{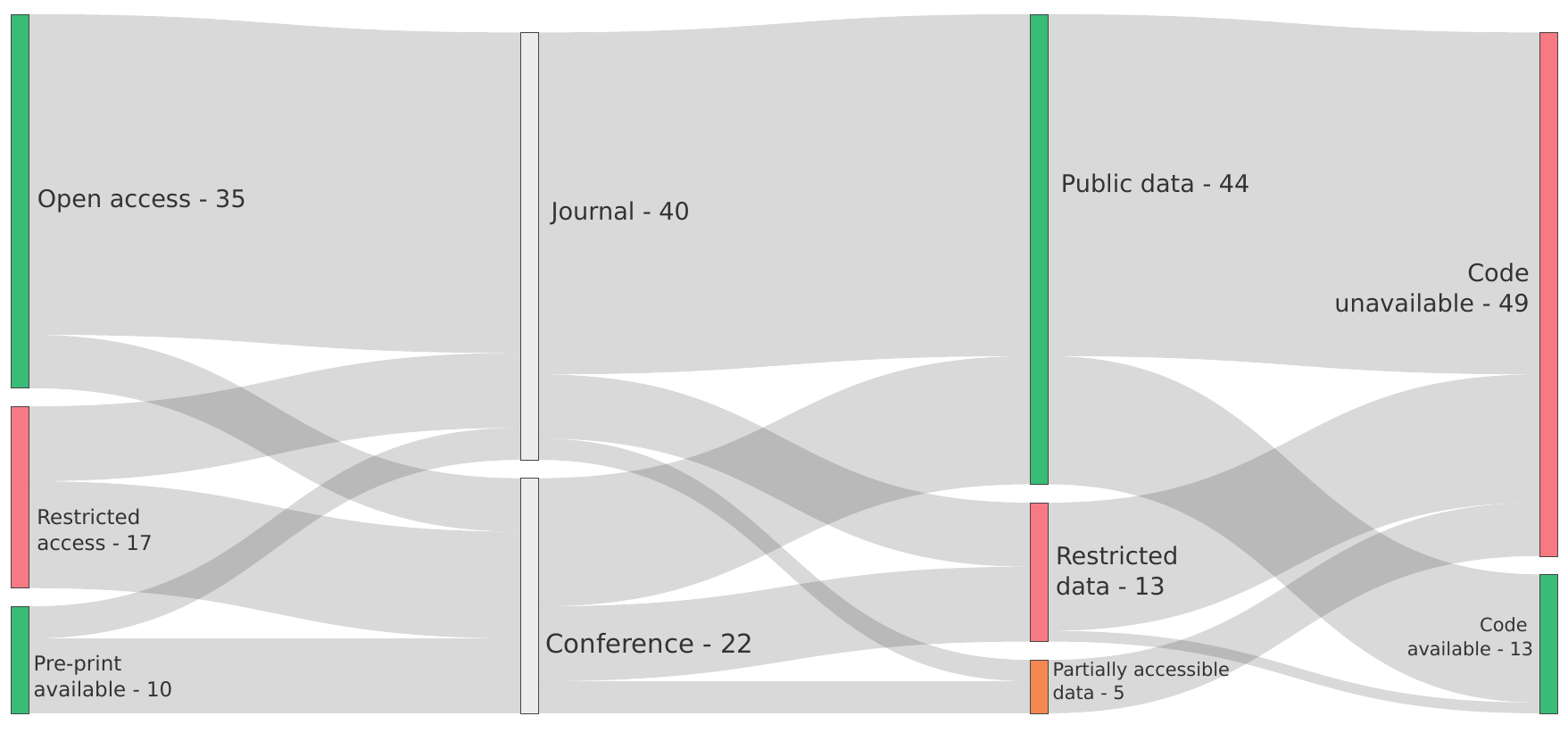}
    \caption{Accessibility of studies. The vertical bars show the frequency of open and restricted access, publication venue, data availability, and source code availability of the selected studies. The gray flows indicate how frequently combinations of characteristics appear together.}
    \label{fig:access}
\end{figure}

We also analyzed the accessibility criteria of papers, verifying data and code availability in the full text, public accessibility of the paper, and publication venue, with results condensed in \autoref{fig:access}. 

We split papers not published as ``open access" in two categories, to understand whether the authors made the full text available as a pre-print before publication. With that in mind, 71\% of the selected studies allows for free access to their methodology and results.

As for the data being readily available from the listed references, some cases included sensitive data on a population (such as call data records), and as such had only partially published data sources alongside the papers, being then labeled as ``partial access to data" in the figure above. As for papers with restrictions on the data sources, most of them either referenced sources no longer available, or that needed some form of authorization or request to the organization responsible for maintaining the data to be accessed.

Lastly, only one fifth of all papers were published alongside their source code, with almost all that do also coming from open access journals. From those, the policies of most of those journals were to encourage authors to also make their code available, being only a requirement at Nature Communications: medicine (for \cite{tab39_Dong2022}), PLOS One (for \cite{tab42_Baquero2018}) and PLOS Neglected tropical diseases (for \cite{tab60_Araujo, tab61_Francisco}).

\subsection{Architectures}

We reviewed the implemented architectures of neural network algorithms from the selected studies, distinguishing among similarly structured neural networks: recurrent neural networks (RNN), convolutional neural networks (CNN), feed-forward artificial neural networks (ANN), and other architectures. We also analyzed hyper-parameters and other modeling choices, such as network size (hidden layers and units), regularization (dropout), and learning rate.

\begin{figure}[!h]
    \centering
    \includegraphics[width=0.65\textwidth]{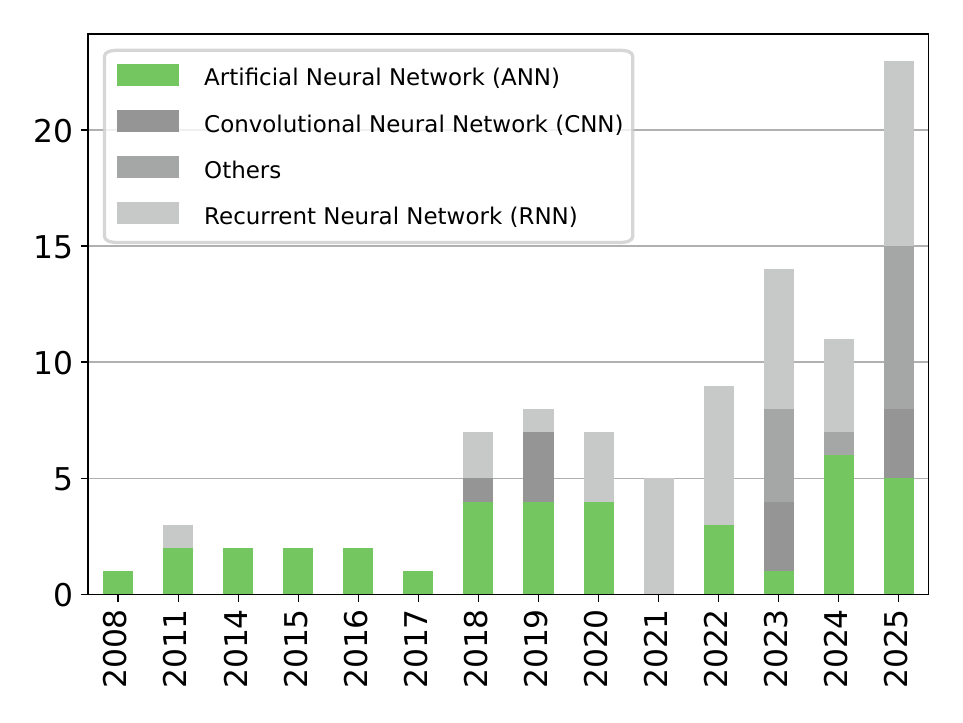}
    \caption{Types of neural network architectures employed by the selected studies and their respective publication year, grouped by structural similarity. The green bars indicate the most commonly used architecture, which was the artificial neural network (ANN).}
    \label{fig:models by year}
\end{figure}


We also compared the optimal hyper-parameters reported in the selected studies for the implemented NNs in \autoref{table:parameters}, describing the parameters through the same architecture grouping criteria. 

    \begin{table}[!h]
    \caption{\label{table:parameters}  Hyper-parameter values by neural network (NN) architecture type. Appearances refers to the number of papers using a neural network architecture that also report a given parameter. As some models define the size of layers or number of hidden units according to the number of lags, namely the number of steps delaying a time series, this information is included here for clarity.}
    \resizebox{1.0\textwidth}{!}{
        \begin{tabular}{llccccc}
        \hline \hline
        \textbf{Architecture category}    & \textbf{Parameter}     & \textbf{Min. value} & \textbf{Max. value} & \textbf{Median} & \textbf{Standard deviation} & \textbf{Appearances} \\ \hline
        \multicolumn{1}{l|}{\textbf{Artificial NN}}& Epochs        & 85         & 8000       & 500    & 2679               & 11          \\
        \multicolumn{1}{l|}{}                   & Hidden layers & 1          & 100        & 1      & 20                 & 30          \\
        \multicolumn{1}{l|}{}                   & Hidden units  & 1          & 256        & 6      & 53                 & 26          \\
        \multicolumn{1}{l|}{}                   & Lags          & 0          & 17         & 4      & 5                  & 10          \\
        \multicolumn{1}{l|}{}                   & Learning rate & 1E-5       & 9E-1       & 3E-1   & 0.3                & 4           \\ \hline
        \multicolumn{1}{l|}{\textbf{Convolutional NN}}& Batch size    & 32         & 100        & 66     & 48.1               & 2           \\
        \multicolumn{1}{l|}{}                   & Dropout       & 0			 & 0.5        & 0.25   & 0.2                & 3           \\
        \multicolumn{1}{l|}{}                   & Epochs        & 100        & 1000       & 550    & 567.4              & 4           \\
        \multicolumn{1}{l|}{}                   & Total layers  & 2          & 161        & 27     & 70.0               & 4           \\ \hline
        \multicolumn{1}{l|}{\textbf{Recurrent NN}}& Batch size    & 10         & 204        & 32     & 57                 & 9           \\
        \multicolumn{1}{l|}{}                   & Dropout       & 0          & 0.8        & 0.25   & 0.22               & 14          \\
        \multicolumn{1}{l|}{}                   & Epochs        & 10         & 8000       & 300    & 1651               & 14          \\
        \multicolumn{1}{l|}{}                   & Hidden layers & 1          & 64         & 3      & 15                 & 15          \\
        \multicolumn{1}{l|}{}                   & Hidden units  & 2          & 1266       & 32     & 344                & 16          \\
        \multicolumn{1}{l|}{}                   & Lags          & 1          & 16         & 9      & 4                  & 5           \\
        \multicolumn{1}{l|}{}                   & Learning rate & 5E-6       & 1E-2       & 2E-3   & 4E-3               & 9           \\ \hline \hline
        \end{tabular}
    }
    \end{table}

Most studies that implemented ANNs do so with the usual fully connected structure, with a few exceptions being given by the NNAR in \cite{tab15_Mustaffa2024, tab16_Chakraborty2019, tab19_Koh2018, tab37_Mahajan2022, tab3_panja2023}, an autoregressive neural network algorithm that relies on statistical criteria for parameter tuning and data processing; and the RBFNN in \cite{tab22_Fadel2020}, which uses radial basis functions as activation functions in the output layer. One study \cite{tab31_Hasanah2020} uniquely implemented a Fuzzy system in order to use clustering for dengue forecast, while also comparing their method to other clustering algorithms.

Two studies featuring CNNs \cite{tab10_Anno2024, tab47_Rehman_2019} implemented a U-Net, which is a convolutional neural network architecture widely used for image segmentation \cite{Ronneberger2015}, to combine satellite images and epidemiologic data to forecast dengue infections in a given region. All others employed the usual convolutional layer stacking of CNNs, with varying number of layers and parameters (see \ref{table:parameters}).

As for the RNNs, the most common architecture used is the long short-term memory (LSTM) NN, appearing in 20 studies. Other less frequent architectures are given by the gated recurrent units (GRU) in \cite{tab2_Ullah2024, tab7_Zhao2023, tab24_Tanvir2021, tab45_Livelo2018} and the BlockRNN \cite{tab3_panja2023}. A Hopfield network is used in \cite{tab23_GARCIAGARALUZ2011} to infer the reproduction number $R_0$ in a Susceptible-Infectious-Recovered (SIR) compartmental dynamic transmission model, with the resulting equation then being integrated and used as a comparison to the epidemiologic data. Other studies \cite{tab47_Rehman_2019, tab59_Lu} also employed neural networks and compartmental models for forecasting.

All mentions of optimizer usage were found in this category, with six studies opting for the Adam optimizer \cite{tab2_Ullah2024, tab24_Tanvir2021, tab25_Zhichao2022, tab33_Li2022, tab34_Kakarla2023, tab42_Baquero2018, tab53_Chen, tab57_Das}, and one each for Adamax \cite{tab34_Kakarla2023} and RMSProp \cite{tab45_Livelo2018}.

From the 13 papers that discuss activation functions, it was found that: rectifier linear units (ReLu) was the most commonly used activation function, found in six studies \cite{tab5_Mussumeci2020, tab7_Zhao2023, tab10_Anno2024, tab28_Ceballos-Arroyo2020, tab42_Baquero2018, tab43_Andersson2019}, followed by the sigmoid activation function \cite{tab14_Anggraeni2018, tab18_Sukama2020, tab32_Dinh2016, tab38_Anggraeni2022}. Studies also used the logistic sigmoid \cite{tab49_Herath2014}, logistic \cite{tab9_Rachata2008}, hyperbolic tangent (tanh) \cite{tab5_Mussumeci2020}, linear \cite{tab32_Dinh2016} activation functions, with the last two used in a layer before the dense layer.

Uniquely mentioned parameters are the Akaike information criterion (AIC) in \cite{tab3_panja2023}, used to select the size of the input layer; the spread factor $\sigma$ for a particle swarm optimization (PSO) algorithm found on \cite{tab35_Anggraeni2024}; momentum on \cite{tab14_Anggraeni2018}, and \cite{tab31_Hasanah2020} made use of fuzzy if-then rules to create input-output pairs for a ANN+fuzzy layer hybrid. Pre-trained embeddings are also used in \cite{tab45_Livelo2018}.

\subsection{Model selection and evaluation}

    \begin{figure}[!h]
        \centering
        \includegraphics[width=1\textwidth]{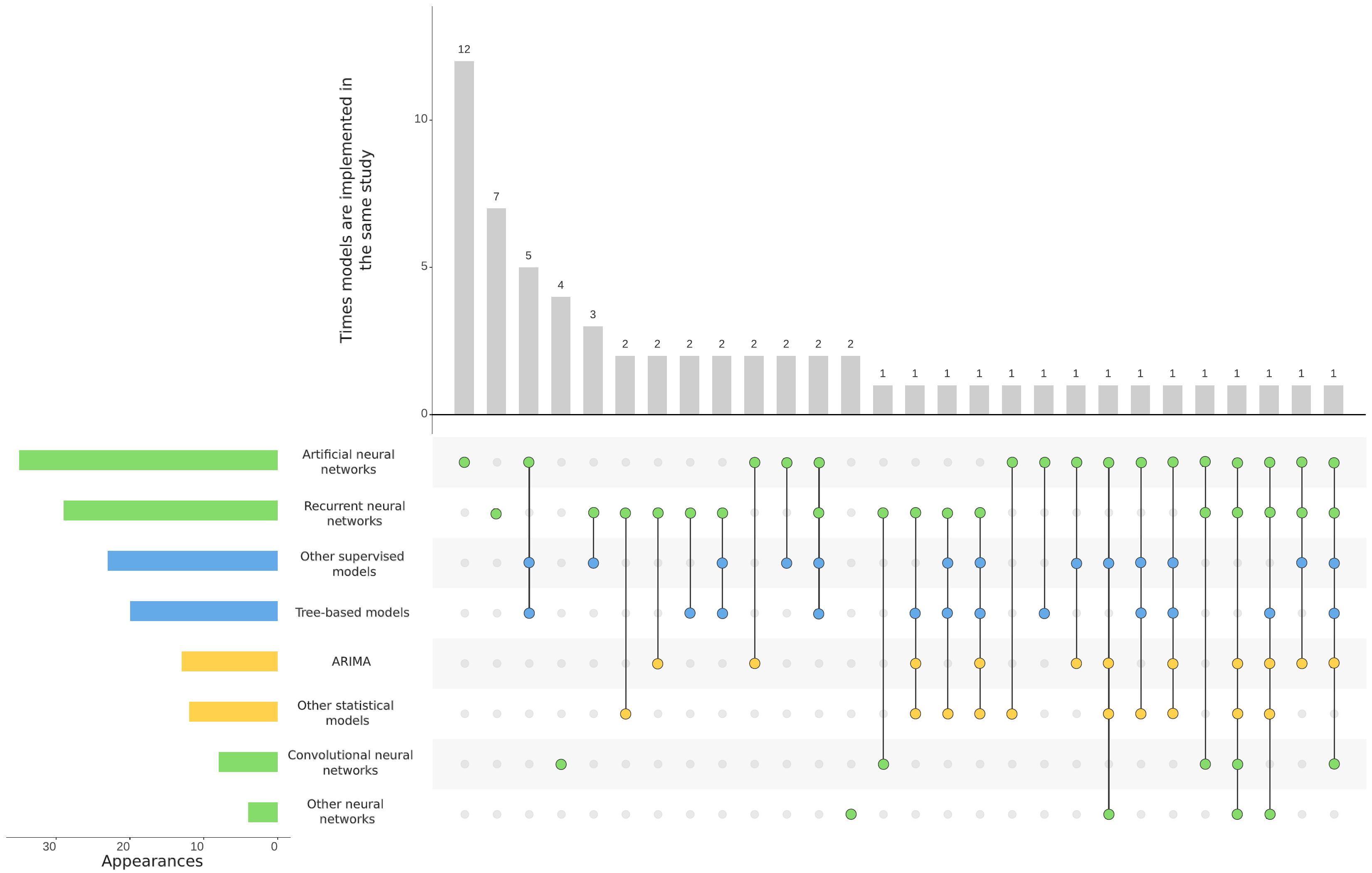}
        \caption{All implemented models found in this review. The horizontal bars accounts for model appearances in the selected studies, with colors grouping structural similarity of these models (green: neural networks, blue: supervised models and yellow: statistical models). Vertical bars show the frequency with which models were used together in a given study.}
        \label{fig:models}
    \end{figure}

Most studies implemented multiple algorithms, comparing  different neural network architectures or machine learning models. ARIMA, being by far the most frequent statistical-based comparator model, was set apart from other similar models, while supervised models were split into tree-based methods, such as XGBoost and Random Forests, and others algorithms. \autoref{fig:dendrogram} shows the variety of NN and comparator models found in this review, alongside their number of appearances. 

Out of the 10 studies that used CNNs, half of them do not employ another comparator model in their report, which is also the highest rate out of all neural network categories (11 out of 34 for ANNs and 6 out of 28 for RNNs). This frequency does not depend on the year of publication of the studies.

When studies did include comparator models, ANNs tended to feature statistical models more often (14 out of 34 entries), with ARIMA being particularly prominent in this case, with 9 papers, while other papers used supervised models for comparisons. RNNs have the opposite distribution, with studies having used supervised models more often and only 10 applying statistical models. It is important to note that including one type of comparator model does not impede the study of also reporting others. Finally, two studies included CNNs with it as a comparator itself: one for an original wavelet neural network architecture \cite{tab3_panja2023}, and another using multiple CNN and RNN architectures \cite{tab7_Zhao2023}. Others compared the CNNs to other neural networks \cite{tab52_Dhaked}, statistical models and tree-based algorithms \cite{tab55_Jaya} and an ensemble of all three model types in \cite{tab62_Salsabiila}. 

Notably, 21 out of all studies included lags, which are delays in a time series set according to a window size defined by the user. They were used in these studies for both epidemic records and the auxiliary variables. Moreover, the 11 studies that also use any type of statistical models as comparators, such as the various autorregressive algorithms shown in \autoref{fig:dendrogram}, favor employing lagged time series to the modeling process, with 65\% of them having this type of feature in the training dataset.

\begin{figure}[htpb]
        \centering
        \includegraphics[width=0.7\textwidth]{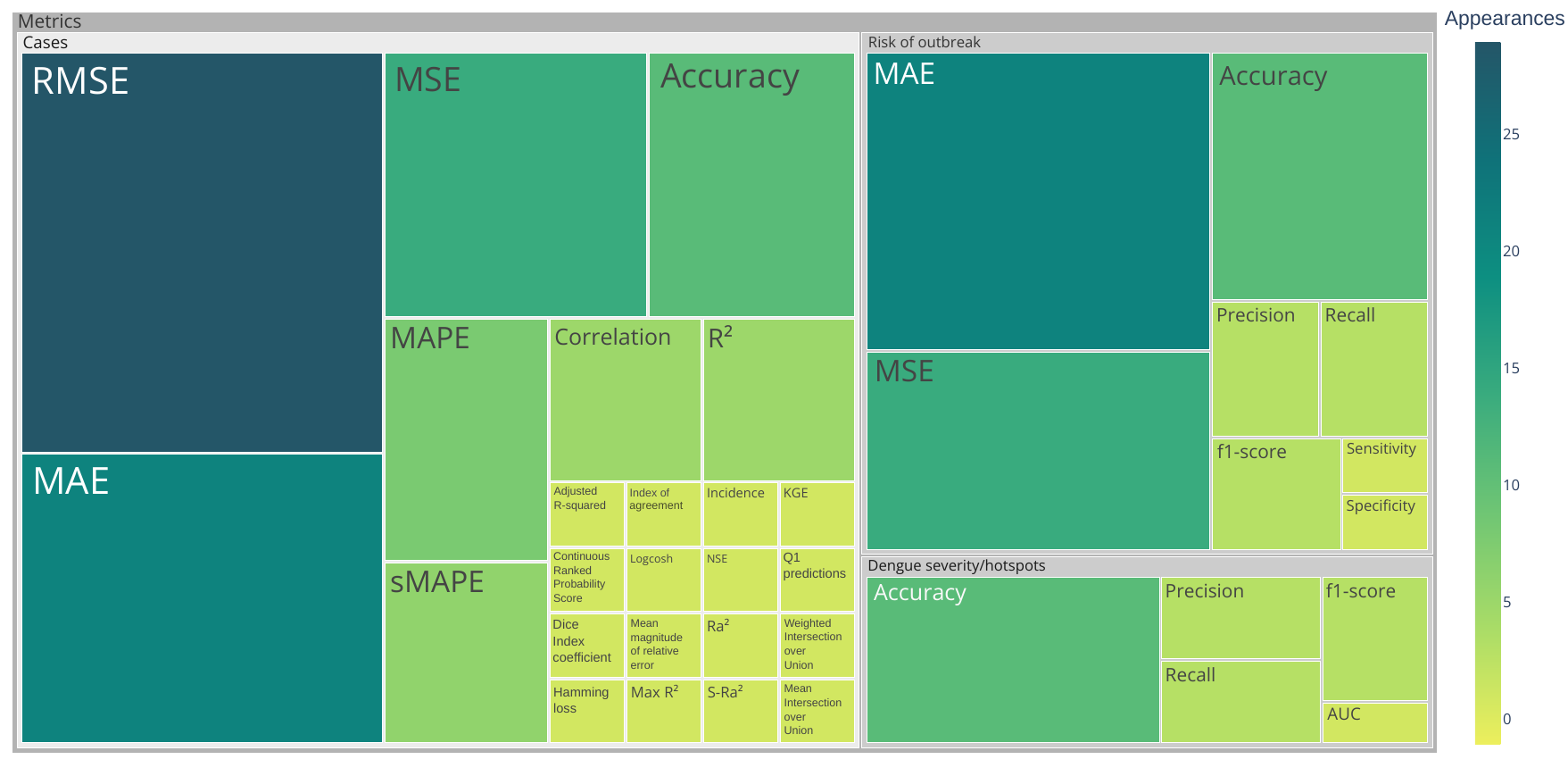}
        \caption{All performance metrics found in the select papers, grouped by target variable: dengue cases forecast, risk of outbreak and dengue severity or hotspots. }
        \label{fig:metrics panel}
    \end{figure}

Within the criteria of study selection, there was significant variation in the studies' specifications, including the formulation of the target feature. Most studies (88\%) predicted the future number of dengue cases based on historical time series. Others made forecasts on the risk of outbreak or dengue severity and its hotspots in a given region. The metrics found throughout this review were grouped accordingly in \autoref{fig:metrics panel}.

Another important distinction between studies is that the forecast window varies depending on the predictive objective and not all studies reported prediction windows: most  of the studies (44\%) made predictions for the whole testing set, which varies in size depending on the data split chosen by authors. When reported, forecasts were mostly made from 1 to 12 week ahead, with the most common forecasting windows employed by these studies ranging from one up to four weeks.

From \autoref{fig:metrics panel}, the most common metric by type of forecast are: Root Mean Squared Error (RMSE) for forecasting dengue fever case counts, Mean Average Error (MAE) for risk of outbreak, and accuracy for detecting dengue severity, or hotspots, in a given area. More than half (58\%) of these studies also employed more than one evaluation metric. For niche metrics, only two studies \cite{tab29_Tanzina2024, tab44_Abeyrathna2019, tab60_Araujo} do not include more frequently used ones when evaluating models through multiple measurements.

    \begin{figure}[!h]
        \centering
        \begin{minipage}{.6\textwidth}
            \includegraphics[width=\textwidth]{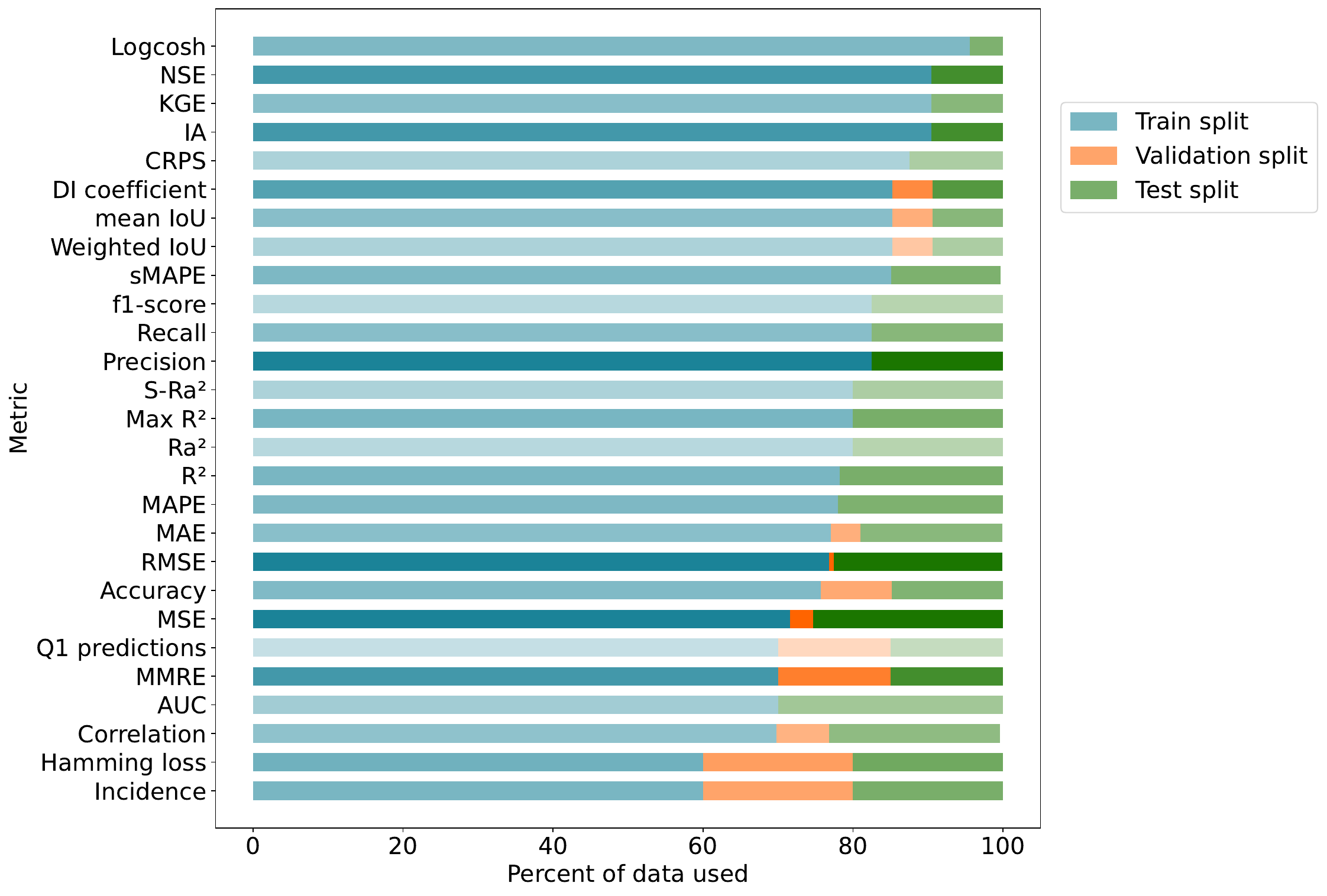}
        \end{minipage}
        \begin{minipage}{.35\textwidth}
            \includegraphics[width=\textwidth]{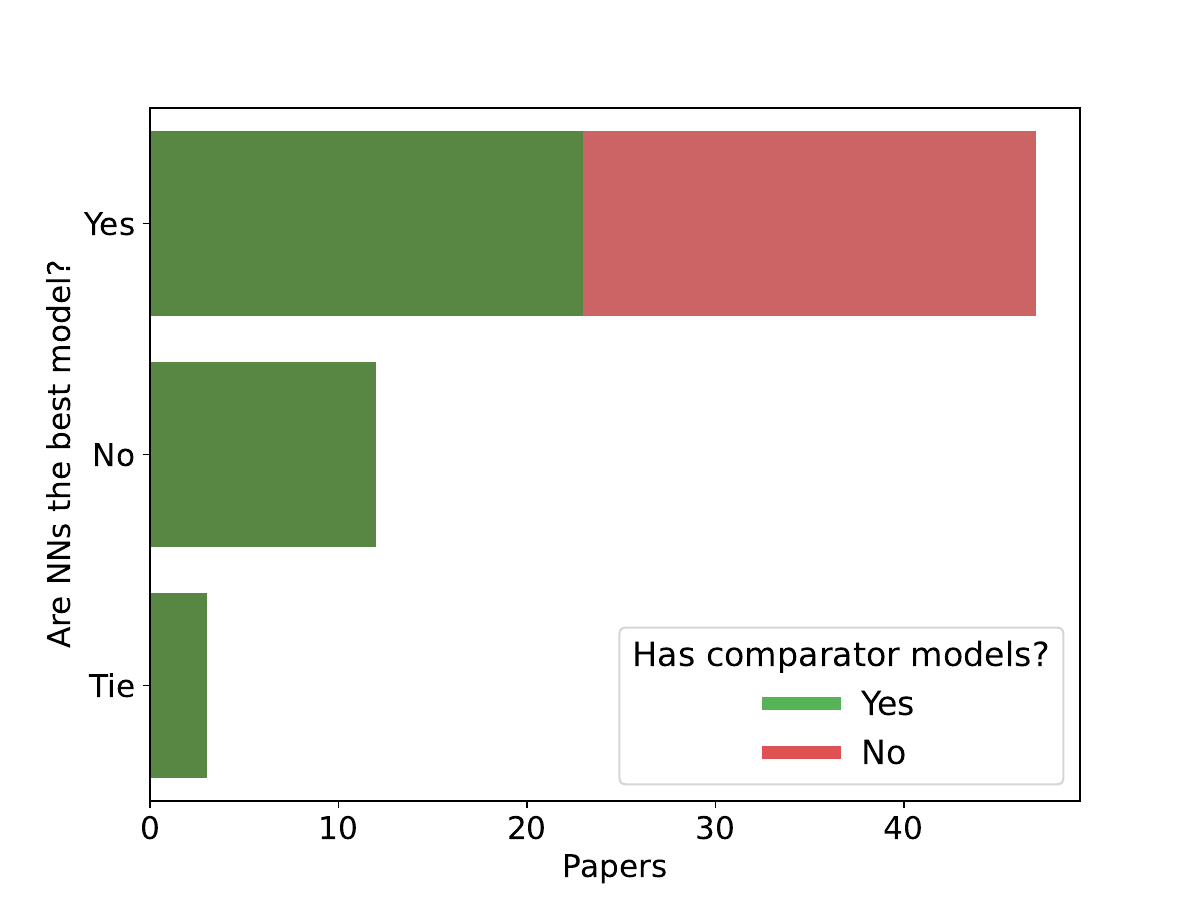}
        \end{minipage}
        \caption{Model evaluation. The left panel shows the averages over the observed splits for each metric, with color gradients indicating the total length of data employed. The right panel shows whether neural networks performed best, and whether the study included comparator models.}
        \label{fig:split and comparators}
    \end{figure}

It was also found that there was not an standardized way of acquiring and treating datasets to train neural networks for dengue forecast, as shown by the different choices in comparison to the type of metric used (see left panel of \autoref{fig:split and comparators}). Cross-validation is also rare, and used specifically in \cite{tab8_xu2020, tab9_Rachata2008, tab12_Souza2022, tab23_GARCIAGARALUZ2011, tab32_Dinh2016, tab44_Abeyrathna2019, tab48_Sousa2015, tab55_Jaya, tab59_Lu}.

It was also noted that some papers only discussed a single architecture and at most compared their results to other similar applications from the literature, without further comparator models implemented, as it can be verified in the right panel of \autoref{fig:split and comparators}.

\section{Discussion}

We reviewed the literature on dengue forecasting using neural networks to analyze trends in methodologies and use of input features, identify research gaps and methodology transparency, and also provide a practical guide for researchers aiming to implement neural networks in their own forecasting work. Most of the studies included in this systematic review reported that their proposed neural network architectures outperformed other machine learning or statistical approaches, suggesting neural networks as a useful method that should be considered alongside other candidate models to forecast dengue incidence, and consequently create strategies that can be later used to guide the decisions of public health agencies. 

Though many risk factors of dengue have been identified, most neural network models limit inputs to meteorological and epidemiological data. Alternative sources of predictor variables leveraged by studies in this review include data from from X (formerly Twitter) \cite{tab5_Mussumeci2020,tab45_Livelo2018, tab62_Salsabiila}, Baidu search data  \cite{Guo2017Oct}, Google street view and satellite images including derived features \cite{tab43_Andersson2019, tab30_Dharmawardana2017, tab47_Rehman_2019, tab58_Hu, tab61_Francisco}, mobile phone records \cite{tab30_Dharmawardana2017}, and mobility data \cite{tab60_Araujo, tab63_Roster2024}. Future studies may evaluate the value of alternative predictors, especially as neural networks can be implemented to handle high-dimensional problems well \cite{Sekmen2024Aug}.

While datasets are expanding, there are still limits to dengue surveillance, including high asymptomatic rates and limits to diagnostic capacity \cite{WHO2024}. One study in this review aimed to address data scarcity using transfer learning\cite{tab8_xu2020}. The authors trained a model on a city with high dengue incidence and later used it to predict the disease in lower-incidence geographies. This was the only study in this review that used transfer learning outside the context of pre-trained CNNs. This may present another avenue for further research, especially in locations where data is scarce, or to tackle under-reporting of dengue cases. Another paper aimed to improve dengue data by addressing the common problem of seasonal fluctuations in case counts, especially at fine temporal and spatial resolution, i.e. zero-inflation, which may lead to inaccurate predictions and performance evaluation \cite{tab61_Francisco}. The authors demonstrated that sparse data interfered with machine learning model accuracy. More research in this space can provide an avenue for a reliable use of data with higher granularity in the future.

There is increasing interest in improving explainability of machine learning models \cite{Linardatos2020Dec}, which is also reflected in this review. For example, SHapley Additive exPlanations (SHAP) were used alongside models to locate climatic drivers of dengue outbreaks \cite{tab53_Chen}.

Changes in climate are expected to greatly impact the global burden of infectious diseases, including dengue \cite{Messina2019Sep,}. While much of the reviewed literature focuses on prediction in unchanged conditions, more work is needed to anticipate dengue outbreaks in places with profound structural changes in climate and socioeconomics, which may disrupt historic patterns of dengue seasonality, the scale and scope of outbreaks, and endemicity. These models may prove critical to anticipating and controlling dengue outbreaks under variable conditions. One study \cite{tab58_Hu} used an ANN to provide global and local projections of dengue distribution and infection considering changes in temporal patterns of climate and socioeconomic data to support policy formulation, however the aforementioned gap in addressing climate change in dengue forecast remains, specially considering that the majority of the studies included in this review make use of meteorological variables. 

We were not able to conduct a meta-analysis of the performance of neural networks in the selected studies, due to limited overlap in datasets and geographies, variations in model architectures, including the train-test split, and lack of open access and transparency in model code and methods. This limits our ability to generalize on the performance of different modeling choices in different contexts, such as epidemic vs. endemic regions of dengue. Future research would benefit from comparing several different model types and architectures on multiple benchmark datasets to better understand how the ideal model varies by their geographic context or data availability.

The field would also benefit from greater transparency and open source coding, as much of the code and data was restricted in the selected studies. One particular study proposed an online and modular platform to automate data retrieval and forecasting for dengue and other arboviruses of interest through several supervised approaches, including multiple possible neural network implementations, which can also be a valuable tool to guide health policies by allowing the benchmark these methods and datasets. \cite{Ganem2024Oct}. Efforts like these can help improve usability of and trust in advanced neural network models by governments to inform dengue control policies.

Neural networks are a promising and increasingly-used tool in dengue forecasting. While much of the field still relies on simple architectures and datasets, several studies have pioneered the inclusion of diverse digital data streams, novel methodologies, such as transfer learning or satellite data. Prediction in the context of changing data generation mechanisms, for example due to climate change, is another promising new avenue of inquiry. More work is needed to compare model performance across geographies and settings and improve the transparency and accessibility of models for policymakers. When implemented carefully, neural networks can aid in the control of dengue. 

\section*{\small{Acknowledgments}}
    Luiza Lober thanks the support given by Fundação de Amparo à Pesquisa do Estado de São Paulo (FAPESP) (grants number 2022/16065-3 and 2013/07375-0). Francisco A. Rodrigues acknowledges CNPq (grant 308162/2023-4) and FAPESP (grants 20/09835-1 and 13/07375-0) for the financial support given for this research. Kirstin Oliveira Roster gratefully acknowledges support from the São Paulo Research Foundation (FAPESP) (grant number 2019/26595-7). 
    
\nocite{*}
\printbibliography


\newpage
\appendix
\section{\label{sec: appendix prisma}The PRISMA flowchart for this study}

\begin{figure}[!h]
    \centering
    \includegraphics[width=0.85\linewidth]{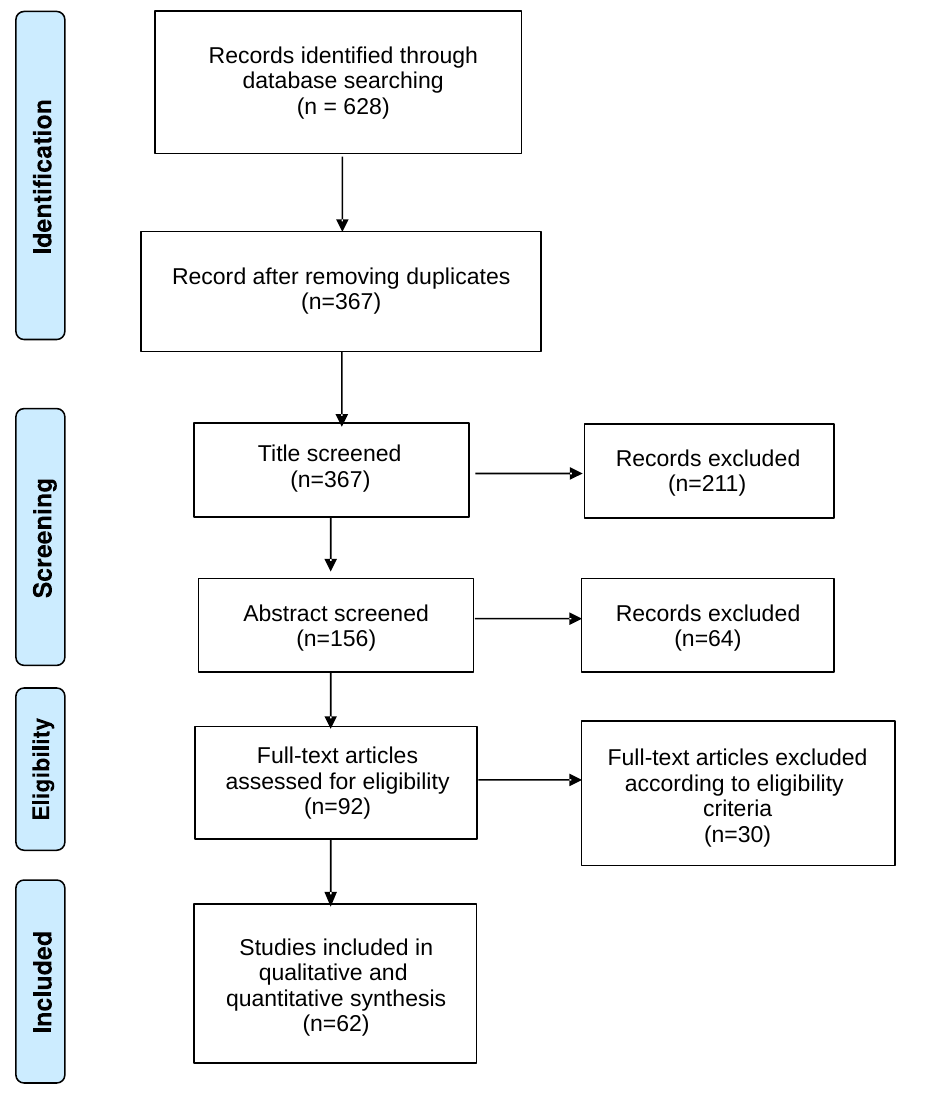}
    \caption{PRISMA \cite{prisma2021} flow chart for this study.}
    \label{fig:prisma flowchart}
\end{figure}

\newpage
\appendix
\section{All NN architectures}

\begin{figure*}[!h]
    \centering
    \begin{subfigure}[t]{0.72\textwidth}
		\includegraphics[page=1, width=\textwidth]{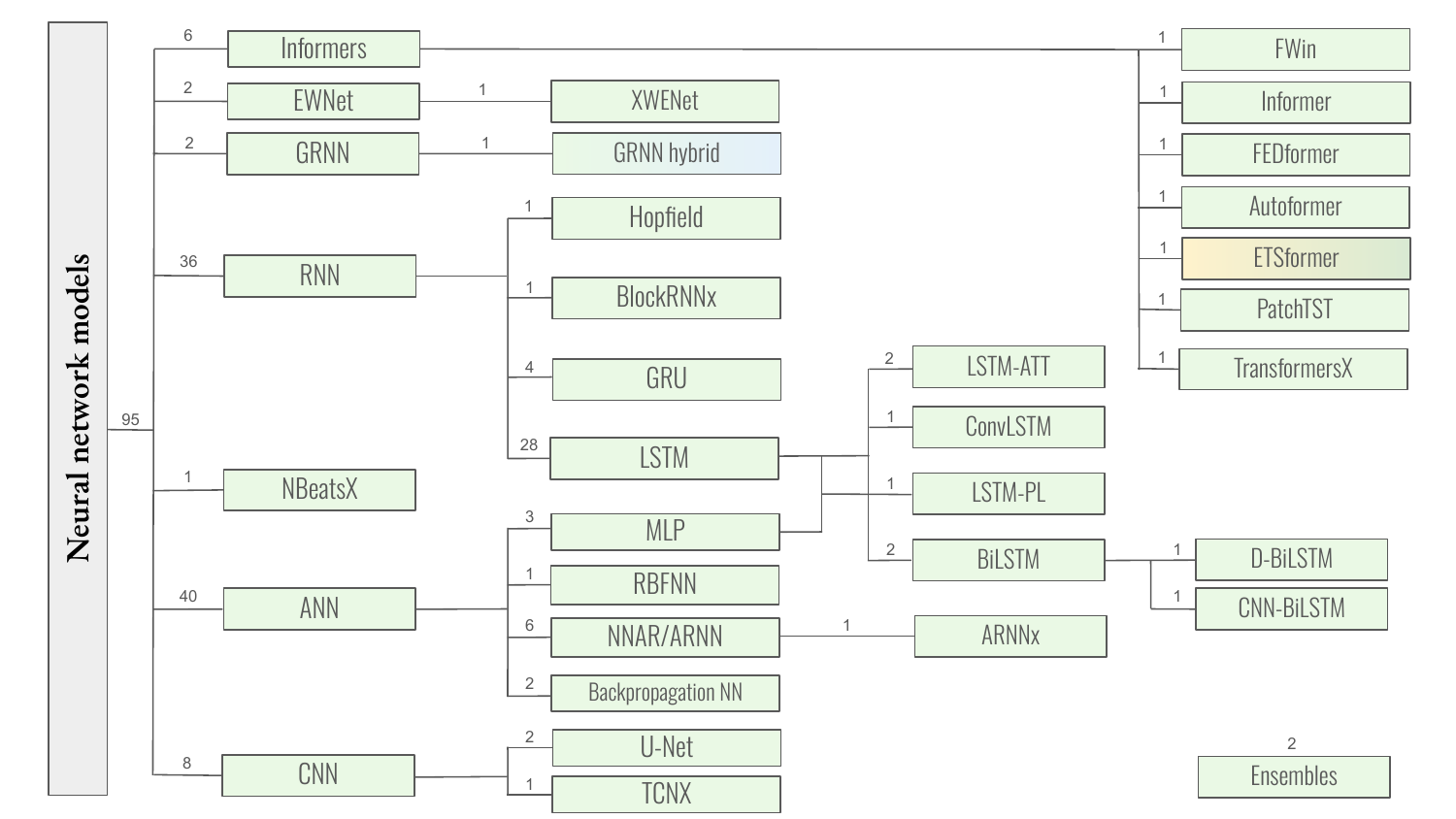}
	\end{subfigure}

    \begin{subfigure}[t]{0.72\textwidth}
		\includegraphics[page=2, width=\textwidth]{dendrogram_numbered.pdf}
	\end{subfigure}

    \begin{subfigure}[t]{0.72\textwidth}
		\includegraphics[page=3, width=\textwidth]{dendrogram_numbered.pdf}
	\end{subfigure}
    \caption{All models found throughout the screening process and included in this review, grouped by structural similarities.}
    \label{fig:dendrogram}
    
\end{figure*}

\end{document}